\documentclass{article}

\usepackage{PRIMEarxiv}

\usepackage[utf8]{inputenc} % allow utf-8 input
\usepackage[T1]{fontenc}    % use 8-bit T1 fonts
\usepackage{hyperref}       % hyperlinks
\usepackage{url}            % simple URL typesetting
\usepackage{booktabs}       % professional-quality tables
\usepackage{amsfonts}       % blackboard math symbols
\usepackage{nicefrac}       % compact symbols for 1/2, etc.
\usepackage{microtype}      % microtypography
\usepackage{lipsum}
\usepackage{fancyhdr}       % header
\usepackage{graphicx}       % graphics
\usepackage{amssymb}
\title{Systemic Fairness
}

\author{
  Arindam Ray \\
  University of South Florida \\
  Tampa, FL, USA\\
  \texttt{arindamray@usf.edu} \\
  \And
    Balaji Padmanabhan \\
  University of South Florida \\
  Tampa, FL, USA\\
    \texttt{bp@usf.edu} \\
   \And
  Lina Bouayad \\
  Florida International University \\
  Miami, FL, USA\\
  \texttt{lbouayad@fiu.edu} 
}

\usepackage{subcaption}
\usepackage{amsmath}

\begin{document}
\maketitle

\begin{abstract}
Machine learning algorithms are increasingly used to make or support decisions in a wide range of settings. With such expansive use there is also growing concern about the fairness of such methods.  Prior literature on algorithmic fairness has extensively addressed risks and in many cases presented approaches to manage some of them. However, most studies have focused on fairness issues that arise from actions taken by a (single) focal decision-maker or agent. In contrast, most real-world systems have many agents that work collectively as part of a larger ecosystem. For example, in a lending scenario, there are multiple lenders who evaluate loans for applicants, along with policymakers and other institutions whose decisions also affect outcomes. Thus, the broader impact of any lending decision of a single decision maker will likely depend on the actions of multiple different agents in the ecosystem. This paper develops formalisms for firm versus systemic fairness, and calls for a greater focus in the algorithmic fairness literature on ecosystem-wide fairness - or more simply \emph{systemic fairness} - in real-world contexts.
\end{abstract}


\section{Introduction} \label{sec:intro}

Machine learning (ML) algorithms are increasingly being used in different applications, from relatively simple ones such as recommending a movie to far more significant ones that can materially impact lives such as helping determine the outcomes of loan applications. The advantages of using algorithms in many of these contexts are clear. Unlike human decision-makers, algorithms can process large amounts of data, do complex computations, scale as the number of decisions taken increases, and do not get fatigued. 
% ADD REFERENCES IN ABOVE PARA on advantage of algorithms over humnas, find papers that make these statements and cite them for the right reason
% ALSO ADD REFERENCES to the two applications we discuss

It is also believed %NEED A REFERENCE % 
that algorithmic decisions are more objective and fair as they are free from human subjectivity and bias. This is important as ML algorithms are extensively used in decision-making in multiple domains and are answering questions such as whether an interviewee will be successful in a job position \cite{CananAlg27:online}, whether a college applicant will be successful and graduate \cite{waters2014grade}, whether a loan applicant will default \cite{baesens2003using} or whether an arrested person might be a repeat offender \cite{berk2017impact}. Predictions driven by ML are clearly affecting significant decisions; it is important therefore that these predictions and the accompanying decisions do not unintentionally hurt some groups of individuals. 

However, in reality, we find counterexamples that question a blind fairness assumption of algorithmic approaches. Correctional Offender Management Profiling for Alternative Sanctions (COMPAS), a tool used by US courts to make parole decisions, was found more likely to assign higher risk scores to African-American offenders than to Caucasians with the same profile \cite{MachineB39:online}. Amazon abandoned a recruitment tool because the algorithm used was discriminating against female candidates \cite{NoSurpri85:online}. In the US mortgage market, otherwise similar Latinx and African-American borrowers pay a 7.9 bps higher rate while borrowing from traditional lenders. FinTech platforms which rely more on algorithms do slightly better, but as we recently saw, still charge 5.3 bps more \cite{bartlett2021consumer} unfortunately. 

Motivated by such issues, there has been growing interest in the literature to better address algorithmic fairness \cite{dwork2012fairness, zemel2013learning,  bolukbasi2016man, hardt2016equality, corbett2017algorithmic, kusner2017counterfactual, buolamwini2018gender, ahsen2019algorithmic, lambrecht2019algorithmic, adomavicius2022integrating}. Most of the fairness literature focuses on classification in static (also called "one-shot" or "batch") settings \cite{hardt2016equality, speicher2018unified} where the complete data is available in advance. Such evaluation on real datasets does have the advantage of being relevant. However, as argued in \cite{d2020fairness} there are long-term fairness questions that cannot be understood and addressed by such static settings. 

To assess the impact on fairness beyond a one-shot or static setting some studies \cite{liu2018delayed, heidari2019long, kannan2019downstream} employ a two-stage model. In these, the first stage decisions are imposed on individuals from two demographic groups, which may drive individuals to take particular actions. The overall impact of this one-step intervention on the entire group is then analyzed in the second stage. There is also work that analytically investigates fairness in dynamic settings beyond two stages including in college admissions \cite{liu2020disparate, mouzannar2019fair}, hiring \cite{liu2020disparate}, lending \cite{wen2021algorithms}, labor markets \cite{hu2018short}, text auto-completion \cite{hashimoto2018fairness} and speech recognition \cite{zhang2019group}. From these studies we know that fairness considerations on decision-making systems need an understanding of the influence on the population at hand, otherwise seemingly fair constraints might in fact amplify the existing differences between groups of the population \cite{mouzannar2019fair}.

As we seek to build on this literature to study long-term dynamics in fairness we also note that most of the current research considers fairness from the point of view of a single focal decision maker, whether it is one college making admission decisions, one organization making hiring decisions or one bank making the lending decision. However, in the real world, none of these institutions work in isolation. Actions of one organization and the outcomes from this are often observed by others in the same industry, who can then take decisions to counter the same and thereby influence the emergent outcomes. We, therefore, need a deeper understanding that takes the perspective of \emph{systemic fairness} rather than just \emph{firm fairness}. Even from a policymaker's perspective, what often matters is the behavior of entire systems rather than individual firms alone. Is fairness at the individual firm level always consistent with fairness at the systemic level and how can we study such issues? This paper presents one approach to doing this by building on the work of \cite{d2020fairness} and using simulations to examine the behavior at the systemic level in the context of lending decisions, chosen as an example. 

% In environments characterized by diverse participants adapting based on outcomes in the environment, there are multiple and repeated interactions of different natures, the outcomes of which are unpredictable. Thus we also argue for the use of a \emph{complex system} lens to study this problem and its \emph{emergent behavior} and to embrace such ideas more broadly as we study systemic fairness considerations in multiple contexts in the real world involving algorithms, humans and an ecosystem of different firms and policymakers. %

Motivated by these considerations in this paper we study a version of the simplified loan setting in prior work \cite{d2020fairness} to mainly illustrate a few key ideas and to make the case for the study of systemic fairness more broadly in the literature. Specifically, we make the following contributions in this paper:
\begin{enumerate}
    \item We present and study \emph{systemic fairness}, a novel and important extension to the fairness literature.
    \item We extend the literature on long-term algorithmic fairness outcomes to a multi-agent system. While there is research done on long-term outcomes, they are focused on single agents currently \cite{d2020fairness} and do not discuss systemic vs firm fairness. 
    \item We develop formalisms to study firm vs systemic fairness and present a framework to study systemic fairness
    \item We present results by studying specific scenarios involving two firms in the loan example and use these to illustrate the differences and importance of considering systemic fairness beyond firm fairness alone.
    \item Once the main ideas are established, we show how the framework can easily be extended to modeling multiple firms and agents using a complex systems lens that focuses on understanding \emph{emergent behavior} and the systemic fairness issues and outcomes arising from this.
%\item We also find that while the individual algorithmic agents may not pursue fairness individually, the system can still be fair. Similarly, while the individual algorithmic agents pursue fairness, the system as a whole can still be unfair.
\end{enumerate}

\section{Background and Literature Review}\label{sec:litreview}

While we focus on algorithmic fairness, the notion of fairness has been studied in psychology, sociology, and philosophy to examine bias and discrimination. Catering to a broad audience the Cambridge Dictionary refers to fairness as ``the quality of treating people equally or in a way that is right or reasonable''. However, measuring this in principle gets quite nuanced, particularly in the context of machine learning algorithms, leading to therefore many definitions of fairness. To help set up the paper, some of these definitions are briefly reproduced here from \cite{mehrabi2021survey} and \cite{verma2018fairness}.

For the formal definitions, we use the following notations, $Y$ is the actual outcome and $\hat{Y}$ is the predicted outcome. We assume the outcome is binary -- $1$ denoting the favorable outcome and $0$ otherwise. $P \in\{1, 0\}$ is the protected attribute, and can be either 1 or 0, depending on whether the individual belongs to a protected group or not.

\textbf{Statistical Parity} \cite{dwork2012fairness} (also known as \textbf{group fairness} \cite{dwork2012fairness}, \textbf{demographic parity} \cite{mehrabi2021survey}): This is satisfied if the likelihood of a positive predicted outcome is same regardless of whether the person is in the protected group or not, i.e., if $P(\hat{Y} = 1 | P = 1) = P(\hat{Y} = 1 | P = 0)$.

\textbf{Conditional Statistical Parity} \cite{corbett2017algorithmic}: The previous definition ignores any other factors which may be the cause for the different outcomes between the groups and hence this definition takes into account a set of legitimate factors L, which can influence the outcome. Based on this definition, an algorithm is fair if the likelihood of a positive predicted outcome is same regardless of whether the person is in the protected group or not after controlling for a set of legitimate factor L, i.e., if $P(\hat{Y} = 1 | L = l, P = 1) = P(\hat{Y} = 1 | L= l, P = 0)$.

\textbf{Predictive Equality} \cite{corbett2017algorithmic}(also known as \textbf{false positive error rate balance}): A classifier algorithm satisfies this definition if the likelihood of a person belonging to the negative class being erroneously predicted to be positive class is the same irrespective of whether the person belongs to the protected class or not, i.e., $P(\hat{Y} = 1 | Y = 0, P = 1) = P(\hat{Y} = 1 | Y= 0, P = 0)$. In other words, the protected and the unprotected group has the same false positive rate (FPR), and thus it is also called false positive error rate balance. A classifier with equal FPR for both groups, will also have equal true negative rate (TNR) for both groups, $P(\hat{Y} = 0 | Y = 0, P = 1) = P(\hat{Y} = 0 | Y= 0, P = 0)$.

\textbf{Equal Opportunity} \cite{hardt2016equality} (also known as \textbf{true positive error rate balance}): A classifier algorithm satisfies this definition if the likelihood of a person belonging to the positive class being correctly assigned to the positive class is the same irrespective of whether the person belongs to the protected class or not, i.e., $P(\hat{Y} = 1 | Y = 1, P = 1) = P(\hat{Y} = 1 | Y= 1, P = 0)$. In other words, the protected and the unprotected group has the same true positive rate (TPR), and thus it is also called true positive error rate balance.
Mathematically, a classifier with equal TPR for both groups, will also have equal false negative rate (FNR) for both groups, $P(\hat{Y} = 0 | Y = 1, P = 1) = P(\hat{Y} = 0 | Y= 1, P = 0)$. For the purpose of this paper, we will use this definition to study systemic fairness, but note that the results will of course depend on which definition is studied. While we show such results for alternative definitions we stress that the importance here is the conceptualization and framework which can generalize broadly to many definitions and use-cases.

The above fairness definitions are examples of \textbf{Group Fairness}, which treats different groups (protected and unprotected) equally. There is another set of fairness definitions, which stems from the idea of giving similar predictions to similar individuals, and
are classified as \textbf{Individual Fairness} \cite{dwork2012fairness, kusner2017counterfactual}.

%We base our research on two streams of literature: a) fairness in machine learning, especially in the sequential scenario and b) complex systems.
%\subsection{Fairness in machine learning in sequential scenario}

There has been a significant body of work that has examined fairness in machine learning. Most of this work has looked at real datasets, and hence are relevant \cite{hardt2016equality, speicher2018unified}. But recent work \cite{d2020fairness} has argued that such work is limited in its ability to study long-term fairness, which necessarily means being able to collect carefully curated longitudinal data with experimental treatments - or instead, approximate those with simulations. Table \ref{table:Lit Review} summarizes the key ideas in the literature as it pertains to looking specifically at longer term dynamics in fairness. %, but, unlike this paper, none of these models the impact of taking actions from predictions on the future data which is used to train models - which is the basis of performative prediction.

\begin{table}[htbp]
\centering
\footnotesize
\begin{tabular}{lllll} \hline 
\textbf{Reference} & \textbf{Horizon} & \textbf{Fairness Notion} & \textbf{Method Deployed} & \textbf{Domain}\\
\hline
          \cite{liu2018delayed} & 2-Stage        &  DP, EO               & Analytical                & Lending\\
          \cite{heidari2019long}&  2-Stage       &  EO               & Analytical     & Social Learning\\ 
          \cite{kannan2019downstream}&  2-Stage
          &  EO, Other          &  Analytical    & College Admission\\ 
          \cite{hu2018short}&  Long Term
          &  Other               & Analytical     & Labor Market\\
          \cite{hashimoto2018fairness}&  Long Term
          & Other           & Analytical     & Text auto-completion\\
          \cite{zhang2019group}&  Long Term
          & DP, EO          & Analytical     & Speech recognition\\
          \cite{mouzannar2019fair}& Long Term
          &  DP              & Analytical     & College Admission \\
          \cite{d2020fairness} & Long Term
          &  EO              & Reinforcement Learning    & Lending, College Admission\\
          \cite{liu2020disparate}& Long Term 
          & EO, PE          & Analytical     & College Admission, Hiring\\
          \cite{wen2021algorithms}&  Long Term
          & DP              & Analytical     & Lending\\ \hline
          \textit{This Paper} & \textit{Long Term} 
          & \textit{EO}          & \textit{Agent Based Modeling }     & \textit{Lending} \\
\hline 
\end{tabular}
\caption{Summary of related work on fairness. 
Note: DP - Demographic Parity, EO - Equality of Opportunity, PE - Predictive equality \cite{verma2018fairness} }
\label{table:Lit Review}
\end{table}

As shown in Table 1, most studies in prior literature use analytical models to assess algorithmic fairness.  In these models, individuals belong to either the protected on the non-protected group.  Each individual attempts to complete a task (such as applying for a job or a university).  A decision maker either approves or denies the individual's request.  Using analytical models, the authors evaluate how the decision-maker's actions impact individuals in the long term.  In the labor market, for example, long-term studies based on analytical models have shown that under statistical parity hiring, job, and wage prospect do not vary significantly with some time lag even if the groups started with different social standing. However, group blind approach and statistically discriminatory hiring (group membership used to infer hidden attributes), contribute to group reputation divergence\cite{hu2018short}.  

Yet, the focus in all of this work is fairness as applied in the context of a single focal firm or decision maker. Such approaches as we noted earlier do not take the emergent outcomes from the actions of multiple entities in the ecosystem into consideration. Our research complements the current studies on long-term fairness and showcases the need of going beyond a single decision-maker to model fairness. The proposed framework can be construed as a test-bed for simulations and analytical methods, to test what-if scenarios and their emergent outcomes. 

%While we examine a simple scenario in this paper for illustration purposes, the main implication here for the community is to develop such approaches for specific problem contexts to gain a deeper understanding of systemic fairness.

\section{Formalism}\label{sec:formal}

In this section we define the key terms we use in this paper to study dynamics related to systemic fairness.

\textbf{Individuals} - Individuals are the people for whom the decisions are being taken. Each individual $k$, where $k \in \mathbb(Z):k \in (0, n)$, has a quality score $Q_k \in [0,1]$, which indicates the probability of success for the individual $k$. The quality is not observable. However, quality is a function of observable features $X_k$ and those are used for making the decision. Individuals also have a protected attribute, $P \in\{0,1\}$, which indicates whether they belong to a protected class or not. The protected attribute cannot be used in making decisions.

\begin{equation}
    Q_{k} = \phi(X_k)
\end{equation}

In a lending scenario, individuals are the borrowers who are seeking credit. Though in the strict definition, borrowers are only those who sought and received credit, we will use the above definition for brevity. If a borrower has a quality score of 0.9, in this context we will take this to mean that probability of repaying the loan is 0.9.

\textbf{Groups} - Groups are collections of individuals based on their protected attributes. The group can be either protected $(P=1)$ or unprotected $(P=0)$.

\textbf{Firms} - Firms are the organizations that take actions from making important classification decisions in the context considered. 
In our example lending scenario, firms are the lending institutions (banks) that lend to the borrowers.

\textbf{Ecosystem} - The ecosystem is the collection of the $m$ firms serving the individuals, the collection of individuals as well as other entities such as policymakers whose actions influence the lending or repayment decisions in any way. For illustration purposes here we consider only the individuals and the firms; however, properly modeling all agents and interactions in a complex systems framework will be important to build realistic models to understand systemic fairness in various contexts.  

\textbf{Decision} - Each individual firm $l$ has its own algorithm to estimate the quality of an individual $k$ based on their observable features $X_k$. We assume the firm has an algorithmic sophistication parameter $s_l$; the higher the sophistication level, the closer is the estimated quality to the actual quality.
\begin{equation}
    \hat{Q_{lk}} = f(X_k, s_l)
\end{equation}
After the estimation of individual quality, the decision of firm $l$ for individual $k$ depends on the firm's risk threshold $\tau_l$ and can be represented by 
\begin{equation}
    D_{lk} = \hat{Q_{lk}} > \tau_l
\end{equation}

In the lending scenario, the decision is whether to approve the loan or not, which is in turn trying to predict the quality of the applicant (whether they will pay back the loan or not).

\textbf{Outcome} - The outcome $O$ is how the decision is affecting the individuals. We distinguish the outcome from the decision, as the outcome may be dependent on one or more decisions.
%Also, the outcome is dependent on the quality of the individual, the same decision can result in different outcomes for two individuals of differing quality.

\begin{equation}
    O_{k} = g(D_{1k}, D_{2k}, ..., D_{mk}) 
\end{equation}

In the lending scenario, it could be a credit score that improves for an individual when she pays back the loan. However, if the loan is not approved the credit score remains unchanged. On the other hand, if the loan was approved for an individual who defaults on the loan, their credit score decreases.

\textbf{Decision Fairness} - There are multiple fairness definitions that are discussed in Section \ref{sec:litreview}, which can be utilized for this. However mainly for illustration we use \textit{equal opportunity} measure as the decision fairness. That is the true positive rate (TPR), the proportion of people predicted positive (i.e., $D_lk$ == 1) when the actual value is positive, is the same for both protected and unprotected groups. The decision fairness is at the firm level. We study the same by analyzing the $TPR Gap$, the difference between the TPR of the protected group and the TPR of the non-protected group.

\textbf{Outcome Fairness} - Similar to the above conceptualization, outcome fairness is observed when the proportion of people with a positive outcome (i.e., $O_k$ == 1) when the actual value is positive, is the same for both protected and unprotected groups. For illustration purposes of systemic fairness, here we take this to also represent the \textbf{Ecosystem Fairness}, as the outcome is based on \emph{all} the firms' decisions in the ecosystem. However, just as fairness is defined in multiple ways, so can metrics for ecosystem fairness which we will leave for future work.

%\textbf{Firm-level Fairness} - We define firm-level fairness, as outcome fairness, considering the decisions and outcomes based on only the firm in consideration.

%\textbf{Ecosystem Fairness} - We define ecosystem fairness, as outcome fairness, considering the decisions and outcomes based on all the firms in the ecosystem.

%\textbf{Fairness Congruence} -

\begin{figure}
    \centering
    \includegraphics[width=3in]{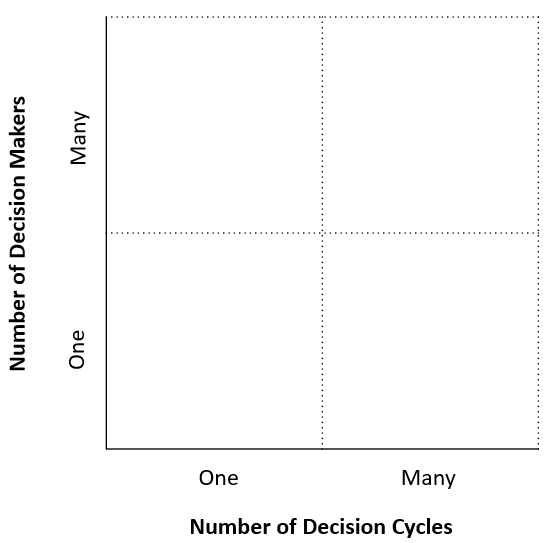}
    \caption{Algorithmic decision-making scenarios}
    \label{fig:2by2}
\end{figure}

We consider four scenarios in a 2 x 2 setting (figure \ref{fig:2by2}). In the horizontal axis, we consider the time horizon - whether the decision-making is done one-time or whether it is a long-term scenario, where the decision must be taken multiple times. The vertical axis, we partition based on whether there is one decision-maker or there is more than one entity taking similar decisions.

The major difference between a snapshot and a long-term scenario is the effect of feedback, where the individual is rewarded (penalized) for their success (failure). For the long-term scenario, we further define these additional terms.

\textbf{Reward} - The reward, ${\Delta Q^+}$ is defined as the increase in $Q_k$ for the individual $k$, once they are successful. For example, in a lending scenario, the individual pays back the loan.

\textbf{Penalty} - The penalty, ${\Delta 
 Q^-}$ is defined as the decrease in $Q_k$, once they is not successful. For example, in a lending scenario, the individual defaults on the loan.

\section{Simulation Study 1}

\subsection{Setup} \label{sec:setup}

To evaluate our framework, we implement an agent-based simulation. We consider a population of 1000 individuals, who belong to a protected or non-protected group with equal probability \footnote{We understand that, in reality, the protected groups are under-represented. However, for ease of exposition we keep the simulation simple first by not considering the additional complexity introduced due to imbalanced classes. However, we have repeated our simulation study 1 with the typically under-represented protected group scenario in Appendix A and find our observations hold true in that scenario.}. Each individual also has a quality score between 0 and 1. Following the setup in \cite{d2020fairness}, we shifted the quality score distribution of the protected group such that the protected group starts out disadvantaged - with the same distribution, but shifted left - compared to the non-protected group. In each time period, a randomly selected set of 100 people apply for a loan to the firms in the ecosystem.

To keep the interpretation as simple as possible here we consider either a single-firm scenario or multi-firm scenario with two firms in the simulation (we subsequently relax this assumption and show results from larger setting as well). We also parameterize each firm by two aspects: \emph{sophistication} and \emph{threshold}. Sophistication indicates the quality (i.e. accuracy or other metric) of the firm's machine learning algorithm for the estimation of the quality of the individual who applied for a loan. The higher the sophistication, the lower the error between the true quality score of the applicant and the estimated quality score. The threshold indicates the risk appetite of the bank. Given the estimate of an individual's quality score, what is the lowest score at which the firm approves the person?

\begin{table}[!htbp]
\footnotesize
\begin{tabular}{p{0.5in}lp{3in}p{1.5in}} 
\hline \hline
Simulation Parameter & Values & Description                  & Link to Formalism    \\ \hline \hline
$n$                      & 1000        & Number of Individuals in the ecosystem    & Individuals                                                          \\
$f$                      & 0.5        & Fraction of Individuals in protected group    & Groups                                                          \\
%p                      & {[}0,1{]}            & Flag indicating whether or not a potential borrowers   belong to a protected group                                                                 & Groups                                                                         \\
$m$                      & 1, 2           & Number of firms in the ecosystem                                                         & Firms)                                                 \\
$t$                      & 50          & Number of Decision cycles                         & Decision Cycles                                                                \\
%Dk                     & {[}0, 1{]}           & The  probability at which each borrower   actually repays the loan                                                                                 & \begin{tabular}[c]{@{}l@{}}Actual Quality \\    \\ (equation 1)\end{tabular}   \\
%Dlk                    & {[}0, 1{]}           & The probability at which the firm expects each borrower to   repay the loan at decision cycle 0                                                    & \begin{tabular}[c]{@{}l@{}}Expected Quality \\    \\ (equation 2)\end{tabular} \\
%Ok                     & {[}0.1{]}            & The probability at which the firm expects each borrower to   repay the loan at decision cycles 1 through 50                                        & \begin{tabular}[c]{@{}l@{}}Outcome  \\    \\ (equation 3)\end{tabular}         \\
$\tau_l$                     & 0.5, 0.6, 0.7, 0.8           & The threshold of quality at which a firm approves an individual     & Attribute of the firm – Risk Tolerance                                         \\
$s_l$                     & 0.7, 0.9   & The noise between the actual quality and the expected quality     & Attribute of the firm – Model Sophistication                                   \\
$\Delta Q^+$                      & +0.05        & Reward - Change in quality score of an individual who succeeds     &  Reward\\
$\Delta Q^-$                      & -0.05        & Penalty - Change in quality score of an individual who fails     & Penalty
%TPR                    & {[}0, 1{]}           & Rate at which the model approved a loan that was actually   repaid                                                                                 & \begin{tabular}[c]{@{}l@{}}Outcome Fairness \\    \\ (equation 4)\end{tabular} \\
%Firm\_TPR              & {[}0, 1{]}           & Rate at which the model of an individual firm approved a   loan that was actually repaid – calculated for the protected and   non-protected groups & Firm-level Fairness                                                            \\
%Ecosystem\_TPR         & {[}0, 1{]}           & Rate at which the models in the ecosystem approved a loan   that was actually repaid - calculated for the protected and non-protected   groups     & Ecosystem Fairness                                                             \\
%TPR\_Gap               & {[}0, 1{]}           & The difference in fairness between the protected and   non-protected group                                                                         & Fairness – can be at the firm or ecosystem level      
\\ \hline \hline                        
\end{tabular}
\caption{Summary of Simulation Parameter Values}
\label{table: SimParam}
\end{table}

Since our context is individuals applying for loans (could be personal loans, credit cards, auto loans etc) and we allow repeated applications from an individual (as in real life) we chose 50 time periods as sufficient to capture a large enough time horizon in a person's life (this time horizon is also sufficient to identify the main qualitative results here). Based on the definition of Equalized Opportunity, we measure the TPR for protected and non-protected groups at the end of each iteration. To eliminate any effect caused by the simulation model's stochasticity, we execute a simulation run for each parameter set 100 times and take the mean of the results. 

For $s$ and $\tau$, though the possible range is between 0 and 1, we take a smaller range of values to more closely reflect practice. For exhaustiveness, we repeat the experiment for the entire range of $s$ and $\tau$ values and noted the results in Appendix C.  

To establish a feedback mechanism, we reward or penalize the person based on her true outcome, and that gets reflected in her quality score. For simplicity, we have kept the reward and penalty of the same magnitude and results in $+0.05$ or $-0.05$ change in her quality score. Table \ref{table: SimParam} summarizes the simulation parameters and values used in our simulation.

%As mentioned previously, in our simulation fairness is represented by the gap in True Positive Rate (TPR) values between the protected and non-protected groups. 

\subsection{Results}\label{sec:results}

Using our simulation model, we investigate the impact of the firm and ecosystem attributes on effectiveness and fairness.  We set up scenarios with fixed model sophistication for a focal firm (FF).  We then vary the approval threshold of FF and look at the impact of different settings for the secondary firm (SF) on the firm and ecosystem fairness levels.  For all the scenarios, we measure 1) the effectiveness of the models at identifying high-quality borrowers (TPR), and 2) The fairness levels (TPR Gap) of these models on the firm and ecosystem fairness levels. 

%The figures below present our simulation results as follows:
 
% - The leftmost column of each figure presents results from having the focal firm as the single firm in the ecosystem.  For example, figure 2c represents the results of having a single firm with a sophistication of 0.7 and an approval threshold of 0.5.

% - The topmost row of each figures presents results from having the secondary firm as the single firm in the ecosystem. For example, figure 2a represents the results of having a single firm with a  sophistication of 0.7 and an approval threshold of 0.7.

% - The cell in the middle of the figure represent the combination of the focal and secondary firms on the edges.  These are the results from having the two firms in the ecosystem (multi-firm).  For example, Figure 2d represents results from an ecosystem where both the focal firm in 2c and the secondary firm in 2a are operating at the same time.

\begin{figure}[!hbp]
\centering
\begin{subfigure}{\textwidth}
    \includegraphics[width = \linewidth]{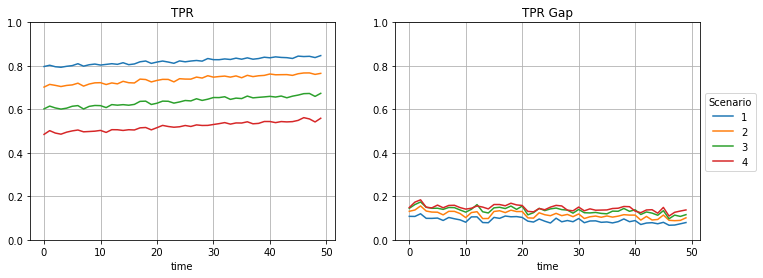}
    \caption{Single firm, focal firm $S = 0.7$ }
    \label{subfig: FF}
\end{subfigure}
\begin{subfigure}{\textwidth}
    \includegraphics[width = \linewidth]{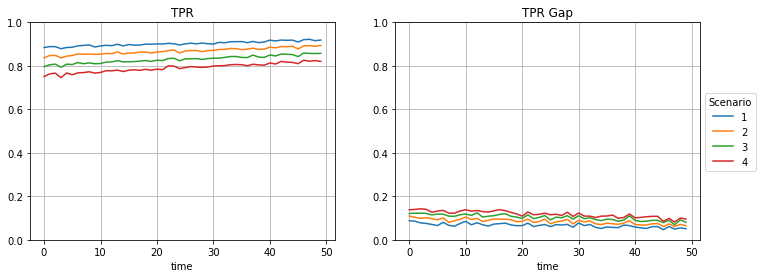}
    \caption{Multi-firm, focal firm $S = 0.7$, second firm $\tau = 0.7, s = 0.7$ }
    \label{subfig: MF1}
\end{subfigure}
\begin{subfigure}{\textwidth}
    \includegraphics[width = \linewidth]{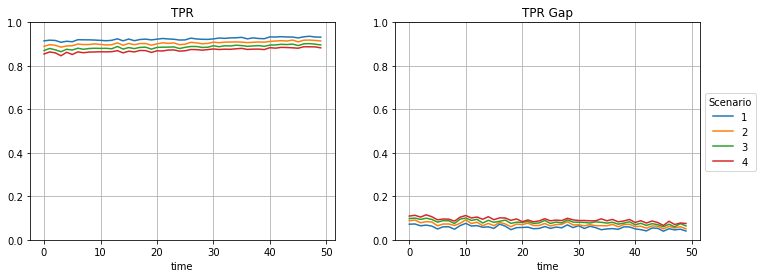}
     \caption{Multi-firm, focal firm $S = 0.7$, second firm $\tau = 0.6, s = 0.9$ }
     \label{subfig: MF2}
\end{subfigure}
\caption{Variation of True Positive Rates and TPR Gap between protected and non-protected groups in Single-Firm and Multi-firm scenarios where the focal firm has low sophistication.
\\
\footnotesize{In all panels, scenario 1, 2, 3, and 4 refers to focal firm's threshold being 0.5, 0.6, 0.7 and 0.8, respectively}}
\label{fig: low_s_ff}
\end{figure}

\begin{figure}[!hbp]
\centering
\begin{subfigure}{\textwidth}
    \includegraphics[width = \linewidth]{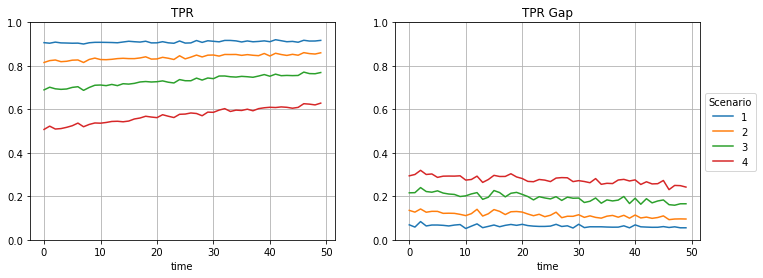}
    \caption{Single firm, focal firm $S = 0.9$ }
    \label{subfig: 09FF}
\end{subfigure}
\begin{subfigure}{\textwidth}
    \includegraphics[width = \linewidth]{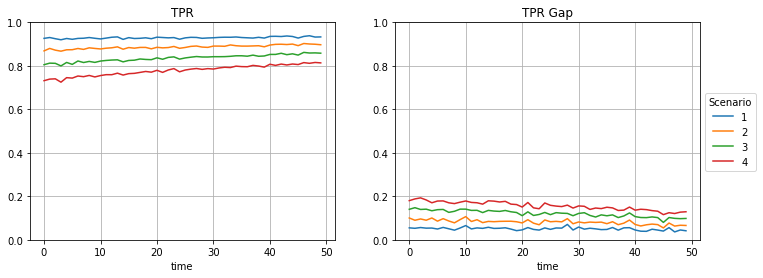}
    \caption{Multi-firm, focal firm $S = 0.9$, second firm $\tau = 0.7, s = 0.7$ }
    \label{subfig: 09MF1}
\end{subfigure}
\begin{subfigure}{\textwidth}
    \includegraphics[width = \linewidth]{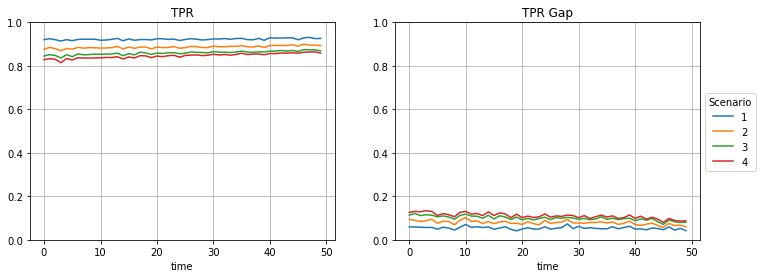}
     \caption{Multi-firm, focal firm $S = 0.9$, second firm $\tau = 0.6, s = 0.9$ }
     \label{subfig: 09MF2}
\end{subfigure}
\caption{Variation of True Positive Rates and TPR Gap between protected and non-protected groups in Single-Firm and Multi-firm scenarios where the focal firm has high sophistication.
\\
\footnotesize{In all panels, scenario 1, 2, 3, and 4 refers to focal firm's threshold being 0.5, 0.6, 0.7 and 0.8, respectively}}
\label{fig: high_s_ff}
\end{figure}

\subsubsection{Model Fairness}

To investigate the impact of the firms' attributes on fairness, we look at the TPR Gap between the protected and non-protected groups in various settings.  

Looking at firm with relatively high model sophistication (Figure \ref{fig: high_s_ff}), we can clearly see the impact of the approval threshold on the firm's fairness level (Figure \ref{subfig: 09FF}).  As the approval threshold of the highly sophisticated FF goes up, the firm will limit applications to very high-quality borrowers.  Given the relatively low-quality levels of individuals within the protected group, this setting will provide more privilege to individuals from the non-protected group. Fortunately, adding a second firm (Figure \ref{subfig: 09MF1} and \ref{subfig: 09MF2}) to this ecosystem offsets the effect of the threshold level and aids in fairness. The second firm, with a lower threshold, gives a chance to individuals from the protected group to get approved. This suggests that in practice when policymakers observe that approval thresholds are very high, they could encourage competition in the market, or even set up institutional loans specifically geared towards individuals from protected groups.

We observe that in the single firm scenario, keeping the sophistication parameter constant, an increase in threshold values results in an increase in the TPR gap between the non-protected and protected groups (Figure \ref{subfig: FF} and \ref{subfig: 09FF}). As the threshold gets higher, the firm is approving a lesser number of people from the protected group, and hence TPR gap is getting higher.  

We observe the same behavior in multi-firm scenarios  (Figure \ref{subfig: MF1} and \ref{subfig: MF2}, as well as in Figure \ref{subfig: 09MF1} and \ref{subfig: 09MF2}). In addition, the ecosystem TPR gap for both the groups in the multi-firm scenario is lower compared to the corresponding TPR gap for each firm in the ecosystem if they were operating alone.  In other words, as we add a new firm to this market, fairness at the ecosystem level also goes up.  Another interesting observation is that, as the approval threshold increases, the TPR gap between the two groups increases. However, this gap is less intense in all multi-firm scenarios compared to its constituent firms in a single-firm scenario.  

While we focus our simulation studies on TPR Gap (EO) as the fairness measure here, we show that the same inferences hold when we consider Statistical Parity Gap as the fairness measure (Appendix B).

\subsubsection{Model Effectiveness}

Maintaining or enhancing fairness is important and is the focus of this work.  However, it is crucial to also evaluate how effective these models are at selecting consumers who will actually replay their loans.  To do that, we track the models' True Positive Rates (TPR) in the scenarios presented above.

When a firm operates as a sole lender in the market, its TPR values get lower as the approval threshold goes up (Figure \ref{subfig: FF}, scenario 1, 2, 3, and 4).  This is mainly due to the fact that, as the firm increases its threshold level, it approves applicants with a tighter range of expected quality levels.  With low model sophistication, the firm is unable to differentiate between high and low-quality applicants within that pool.  

When a secondary firm (Figure \ref{subfig: MF1} and \ref{subfig: MF2}) is added to this ecosystem, however, we observe a significant impact on the models' effectiveness.  Overall, adding a firm with equal or higher sophistication and a higher approval threshold to the market improves the TPR values. It can be explained by the fact that by going through more than one application process (with approval thresholds relatively high), the combined models act as an ensemble and are better able to select/approve high quality applicants from both groups.

In practice, this implies that, when individual firms are observed to have less than optimal models, policymakers should encourage the increase of the number of firms in the market; as this will improve the ability of qualifying individuals from both the protected and non-protected groups to get approved.

%Looking at the entire ecosystem, policy makers can not only help provide a fair treatment for individuals from the protected group, but also help improve the firms' identification of high quality consumers.

\subsubsection{The Longitudinal Effects}
In all the scenarios above, we track the model fairness and effectiveness levels at multiple rounds of loan applications.  This allowed us to investigate the long-term effects.  In our simulation settings, we observe that the model effectiveness (TPR) increases over time.  We also see a steady decrease in TPR gaps, indicating higher fairness over time.  This can be explained by the fact that borrowers that repay their loan are rewarded for their positive behavior by enhancing their quality scores and vice versa.  With these reward and penalty effects, the gap between high and low-quality applicants gets wider over time; making it easier for models to detect and approve high-quality applicants.  

Overall, we can conclude that in the settings that we studied above, the multi-firm scenario not only offers better TPR but also a reduced TPR gap.  This indicates that, at the ecosystem level, the two groups are treated more equally, and more generally, illustrates the value of taking a systemic fairness lens to such important settings. 

\section{Simulation Study 2}

\subsection{Setup}
In the previous simulation study we assumed the firms have the same sophistication and threshold for both the protected group and the non-protected group. We relax those assumptions in this study.

\subsection{Results}

\subsubsection{Implicit Bias} 
\begin{figure}

    \begin{subfigure}{\textwidth}
        \centering
        \includegraphics[width=\linewidth]{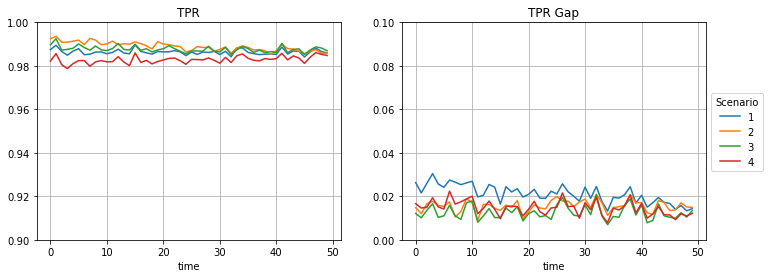}
        \caption{Firms with low thresholds (0.3)}
        \label{subfig:unintentional_bias_0.3}
    \end{subfigure}

    \begin{subfigure}{\textwidth}
        \centering
        \includegraphics[width=\linewidth]{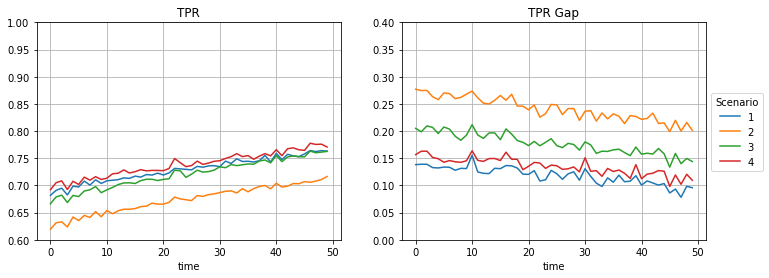}
        \caption{Firms with high thresholds (0.8)}
        \label{subfig:unintentional_bias_0.8}
    \end{subfigure}
    
\caption{Effect of implicit bias exhibited by firms.}
\label{fig:unintentional_bias}
\end{figure}

This is a scenario where the firms are able to assess a member of non-protected group better than someone from a protected group. This could be due to a lack of data on the protected group or historical bias within such data. In our environment, we simulate it by firms having different sophistication levels for the protected group and the non-protected groups. For example, in Figure \ref{subfig:unintentional_bias_0.3} scenario 1 both the firms have higher sophistication ($s = 0.9$) for the non-protected group compared to the protected group ($s = 0.7$). 

However, in case of a regulated industry such a scenario may not be acceptable to the industry regulator / policymaker and it may advice the firms to have the same sophistication level for the two groups. Given this mandate, the firms may have three possible actions - a) improve the sophistication level for the protected group to match it to that for the unprotected group (Figure \ref{subfig:unintentional_bias_0.3} Scenario 2, $s = 0.9$ for both groups), b) lower the sophistication level for the unprotected group to match it to that for the protected group (Figure \ref{subfig:unintentional_bias_0.3} Scenario 4, $s = 0.7$ for both groups), or c) meet somewhere in between (Figure \ref{subfig:unintentional_bias_0.3} Scenario 3, $s = 0.8$ for both groups).

In an ideal scenario, which is depicted in scenario 2, the firms will have higher sophistication ($s = 0.9$) for both groups, which results in better efficiency (higher TPR) as well as better fairness (lower TPR gap). However, in reality, it may be harder to achieve and firms may opt for scenario 3 or 4. In both scenarios 3 and 4, the fairness was better (lower TPR Gap), than in scenario 1, and closer to scenario 2. The efficiency, TPR, drops as the sophistication drops. Scenario 4 is worse even than scenario 1, where at least the non-protected group has a better sophistication level.

However, the results above vary depending on the threshold level of the firm. In figure \ref{subfig:unintentional_bias_0.8}, where the threshold for the approval is high ($\tau = 0.8$), a high sophistication actually has a reverse effect. We can see that scenario 2 in this case yields the worst efficiency (lowest TPR) and worst fairness (highest TPR gap). This observation can be explained by the fact that in this case, the threshold being very high only a small portion of the good-quality people gets approved. A lower sophistication allows for a larger variation in estimation of the quality and hence more people get approved, who still have reasonably good quality. Thus in high threshold firms, relaxing the sophistication leads to higher efficiency and fairness.

\subsubsection{Explicit Bias} Though we do not think this may be practiced in real life, we evaluate this scenario from purely an academic perspective. We design this by setting a different threshold for the protected group, compared to the non-protected group for the firms. 

\begin{figure}
    \centering
    \includegraphics[width=\linewidth]{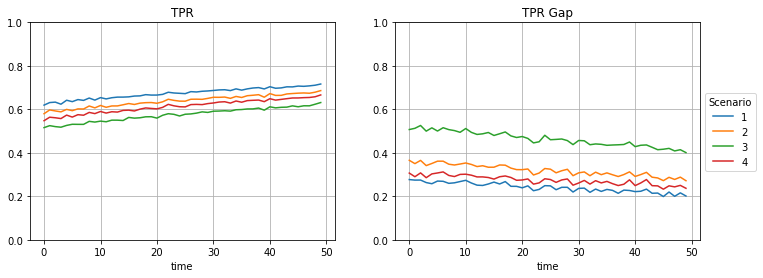}
    \caption{Effect of Explicit bias exhibited by firms. Scenario 1, both firms fair, have threshold value 0.8 for both the protected and non-protected groups. Scenario 2, firm 1 is unfair, have higher threshold value 0.9 for the protected groups.
    Scenario 3, both firms are unfair, have higher threshold value 0.9 for the protected groups.
    Scenario 4, firm 1 has higher threshold value for protected group, firm 2 has higher threshold value for the non-protected group.}
    \label{fig:intentional_bias}
\end{figure}

We observe the results in Figure \ref{fig:intentional_bias}. In scenario 1, both firms are fair, that is the threshold for both firms are same for the two groups. In scenario 2, the threshold for the protected group is higher in firm 1, i.e. the protected group is evaluated for a higher standard. This leads to having lesser number of people from protected group being approved for scenario 2, and thus the TPR gap increases. In the third scenario, both banks are unfair towards the protected group, which results in further widening of the TPR Gap. In scenario 4, one firm is unfair towards protected group and the other firm is unfair towards the non-protected group, which results in TPR gap being smaller.

\section{Simulation Study 3}

\subsection{Setup}
To simulate more realistic scenarios, in this study we increase the number of firms from $2$ to $20$  and the number of individuals from $1000$ to $100,000$. We also relax the assumption that all individuals apply for loans at all the firms. Typically as there is a direct or indirect cost involved with each application, individuals may not be willing to apply to all firms. For example, while shopping for a loan may land a better interest rate, each application has an indirect cost of the hard query into the applicant's credit report which reduces the applicant's credit score. 

\subsection{Results}
To better assess the fairness levels in realistic settings, we first measure the model fairness and effectiveness at the baseline (scenario 1), where all individuals apply to all $20$ firms. In this case, we observe high TPR levels and low TPR Gaps.  This is because most of the people get approved by one firm or anoother in this larger ecosystem. 
In scenario 2, people from the non-protected group apply to all $20$ firms, whereas people from the protected group only apply to three, randomly selected, firms. As the number of firms came down, the approval rate also came down, and hence the TPR value. As the applications are fewer for the protected group, they are more affected compared to the non-protected group and hence the TPR Gap also increases.
In Scenario 3, we simulate the situation such that the $3$ firms the protected groups are choosing are not randomly among all $20$ firms in the ecosystem, but from half of them whose risk thresholds are lower. We see an improvement from scenario 2, in both TPR and TPR Gap, as choosing the lower threshold firms we increase the approvals for the protected group, and hence TPR improves, and TPR Gap decreases.
In scenario 4, we also simulate that a firm with lower threshold might have a higher cost of service, which might increase the chance of a negative outcome. For example in lending, a bank with a lower risk threshold may have high-interest rate, which in turn increases the probability of default. We observe that there is a slight drop in TPR and also TPR Gap higher compared to Scenario 3.
From these scenarios, we can infer that applying to as many firms as possible helps in improving fairness, and the policymaker may introduce rules that eliminate the application cost barriers. In absence of such rules, it may be best to nudge the individuals to a specific type of firm than others, even if their service cost is higher.  

\begin{figure}
\centering
    \includegraphics[width=\linewidth]{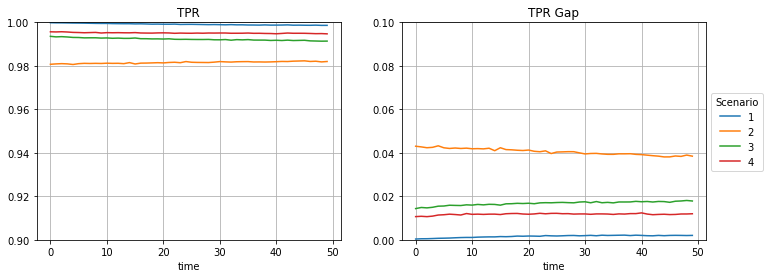}
   
\caption{Effect of selective applications by individuals.}
\label{fig:selective_applications}
\end{figure}

\section{Discussion}
From natural language processing models to recommender systems, algorithms have been investigated by researchers to detect and measure biases toward protected groups \cite{hutchinson2020social, ekstrand2022fairness, mansoury2020feedback}.  This stream of research has been invaluable in improving our understanding of how algorithms that support our decision-making processes are affecting individuals in virtually all domains. Motivated by this, we have seen a new stream of research looking into developing methods to curb these biases \cite{mansoury2021graph, wen2021algorithms}.  

The magnitude of the algorithmic fairness problem has also prompted policymakers to develop legislation aiming at limiting the power of such algorithms in the practice \cite{green2022flaws}.  This is currently being implemented through the emphasis on human discretion; reducing the automation of decision-making.  While these solutions can help resolve the problem in the short run, these methods can 1) increase the cost associated with using and maintaining the algorithms in the practice, 2) reduce the benefits of deploying these algorithms, and 3) introduce human biases into the decision-making process.

Given the significance of these issues in practice, it is essential that our models and frameworks take into consideration the entire ecosystem at play in various scenarios. If not, seemingly good solutions, when deployed in practice, may fall short of the fairness outcomes we might expect. With this in mind, we extend the fairness body of literature by developing and evaluating a fairness framework for systemic fairness at the ecosystem level and argue for more work in this direction.  Such settings can provide valuable insights that can guide both individual firm decision-making as well as policy initiatives. For instance, in the simple setting we examined here, we find through the systemic fairness lens that lower levels of fairness at a specific firm can be offset by the establishment of other firms in the market.  

Expanding the frame of investigating fairness to the systemic level, opens the door for several streams of research.  As a future step for our study, we plan to introduce more heterogeneous firms in the ecosystem and evaluate their effect on both firm and systemic fairness.  We also investigate the effectiveness of different interventions developed in prior literature, at the individual firm level, in the ecosystem setting.  While this study has successfully used the lending domain as a test bed for the systemic fairness framework, more studies are needed to evaluate the same framework in other domains.

%that the relationship of TPR and TPR gap with sophistication is not monotonous, but depends on the corresponding threshold value. For example, with the threshold of 0.5, TPR for both the non-protected group and protected group drops with the drop in sophistication level. However, with the threshold value of 0.8, such a drop in sophistication level shows a drop in TPR value for the non-protected group and an improvement of TPR value for the protected group.
\newpage

\appendix
\section{Under-represented protected group}
In the paper, we have assumed that the population is equally distributed between the protected and the non-protected groups. Though in reality, the protected group is typically underrepresented. For example, the U.S. Census Bureau estimates 13.6\% of population is African American and 18.9\% is Hispanic or Latino \cite{USCensus78:online}. We rerun our simulation study 1, with the protected group to be the same fraction of the population, and we find that our observations hold true with the under-represented protected group.

\begin{figure}[!hbp]
\centering
\begin{subfigure}{\textwidth}
    \includegraphics[width = \linewidth]{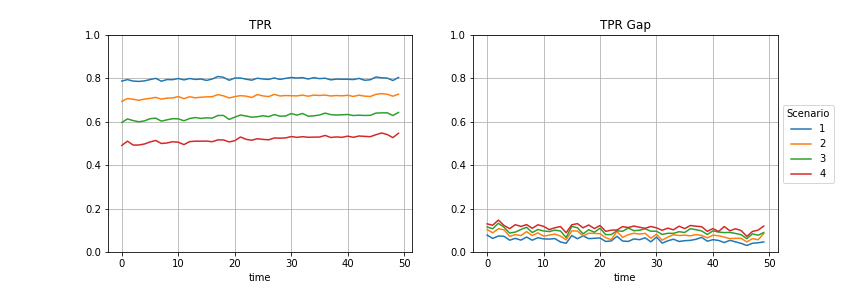}
    \caption{Single firm, focal firm $S = 0.7$ }
    \label{subfig: FF_A}
\end{subfigure}
\begin{subfigure}{\textwidth}
    \includegraphics[width = \linewidth]{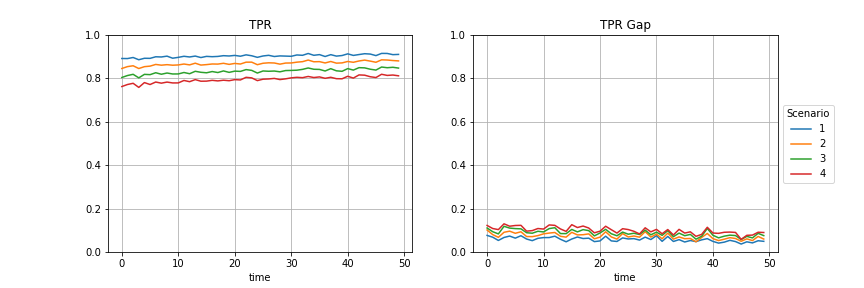}
    \caption{Multi-firm, focal firm $S = 0.7$, second firm $\tau = 0.7, s = 0.7$ }
    \label{subfig: MF1_A}
\end{subfigure}
\begin{subfigure}{\textwidth}
    \includegraphics[width = \linewidth]{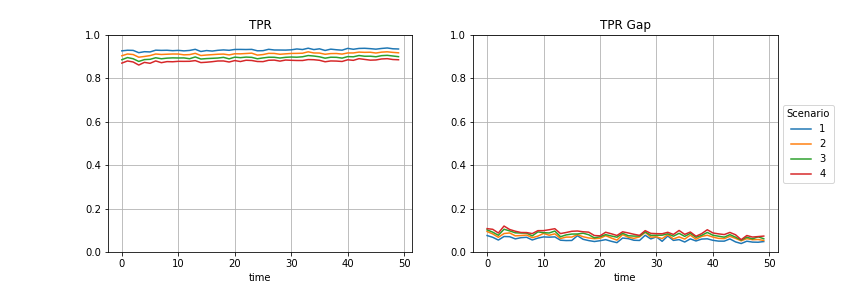}
     \caption{Multi-firm, focal firm $S = 0.7$, second firm $\tau = 0.6, s = 0.9$ }
     \label{subfig: MF2_A}
\end{subfigure}
\caption{Variation of True Positive Rates and TPR Gap between protected and non-protected groups in Single-Firm and Multi-firm scenarios where the focal firm has low sophistication.
\\
\footnotesize{In all panels, scenario 1, 2, 3, and 4 refers to focal firm's threshold being 0.5, 0.6, 0.7 and 0.8, respectively}}
\label{fig: low_s_ff_A}
\end{figure}

\section{Generalizability of the results for other fairness measures}

In our simulation study 1, we also measure the Statistical Parity gap, as an additional fairness measure. We observe the results that we discuss with TPR Gap also hold true for the Statistical Parity Gap measure. This observation helps to establish that the systemic fairness phenomenon is generalizable over multiple fairness measures, and not just unique to the EO measure.

\begin{figure}[!hbp]
\centering
\begin{subfigure}{\textwidth}
    \includegraphics[width = \linewidth]{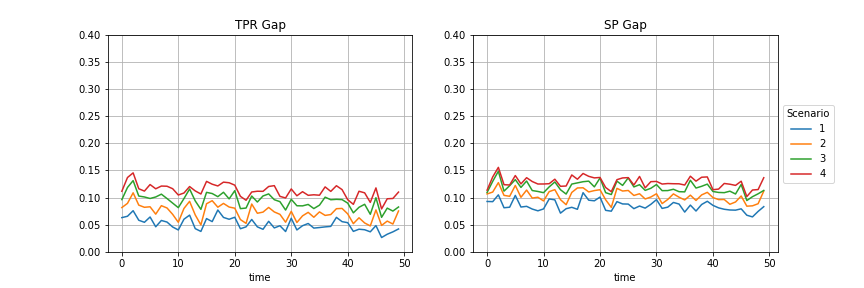}
    \caption{Single firm, focal firm $S = 0.7$ }
    \label{subfig: FF_B}
\end{subfigure}
\begin{subfigure}{\textwidth}
    \includegraphics[width = \linewidth]{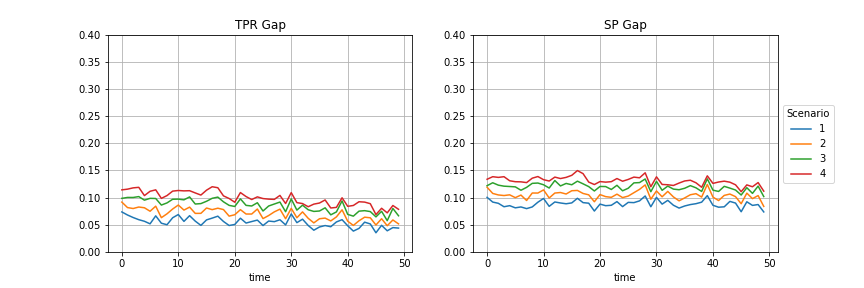}
    \caption{Multi-firm, focal firm $S = 0.7$, second firm $\tau = 0.7, s = 0.7$ }
    \label{subfig: MF1_B}
\end{subfigure}
\begin{subfigure}{\textwidth}
    \includegraphics[width = \linewidth]{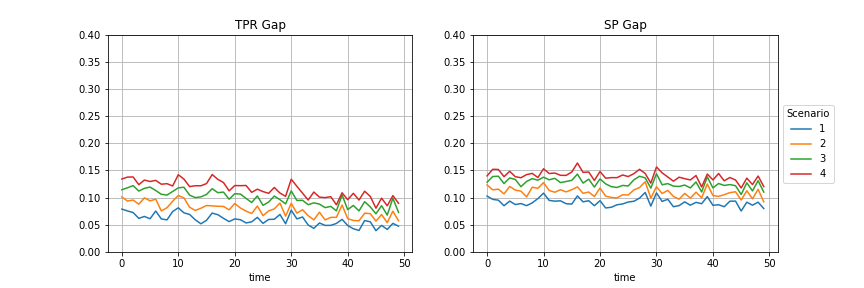}
     \caption{Multi-firm, focal firm $S = 0.7$, second firm $\tau = 0.6, s = 0.9$ }
     \label{subfig: MF2_B}
\end{subfigure}
\caption{Variation of TPR Gap and SP Gap between protected and non-protected groups in Single-Firm and Multi-firm scenarios where the focal firm has low sophistication.
\\
\footnotesize{In all panels, scenario 1, 2, 3, and 4 refers to focal firm's threshold being 0.5, 0.6, 0.7 and 0.8, respectively}}
\label{fig: low_s_ff_B}
\end{figure}

\section{Model effectiveness and fairness on a broader range of threshold and sophistication values}

While both the sophistication and threshold parameters can take any value between 0 and 1, in the simulation studies we have used the values in moderation, typically between 0.5 and 0.9. A threshold value of 0.5 for a bank indicates that the bank will approve an individual whose estimated probability of payback is 0.5. With that definition in mind, we think that it will be highly unlikely for the banks to have a threshold below that point. Similarly, a sophistication of 0.5 indicates that the error of estimating the quality of the individual has a variance of 0.5. For a measure, like quality, whose value is between 0 and 1, it will again be highly unlikely the sophistication of the algorithm be less than 0.5. 

\begin{figure}[!hbp]
\centering
\begin{subfigure}{\textwidth}
    \includegraphics[width = \linewidth]{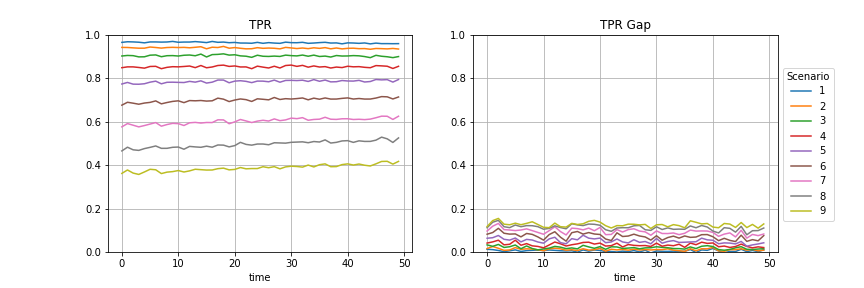}
    \caption{Single firm, $S = 0.7$ with $\tau$ varying from 0.1 to 0.9 in scenario 1 to 9 }
    \label{subfig: FFT_C}
\end{subfigure}
\begin{subfigure}{\textwidth}
    \includegraphics[width = \linewidth]{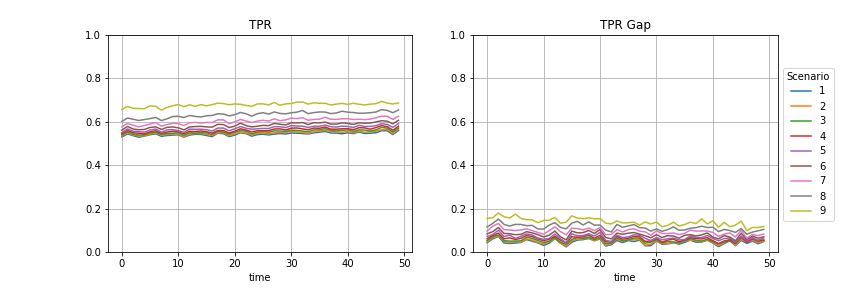}
    \caption{Single firm, $\tau = 0.7$, with $s$ varying from 0.1 to 0.9 in scenario 1 to 9  }
    \label{subfig: FFS_C}
\end{subfigure}
\caption{Variation of TPR and TPR Gap between protected and non-protected groups in Single-Firm scenarios with varying thresholds and sophistication levels.}
\label{fig: low_s_ff_C}
\end{figure}

However, for the exhaustiveness of analysis, we repeat the simulation study 1a, for the threshold and sophistication values covering the entire possible range, and observe the trends continue for the entire range. Figure \ref{subfig: FFT_C}, indicates that the effect we have seen in the simulation study 1 that as the threshold is reduced the efficiency improves as well as fairness improves holds true for the entire range of threshold values. Figure \ref{subfig: FFT_S} indicates as the sophistication level increases, the TPR improves, however TPR gaps also increases as it the bank is able to predict individual quality with better accuracy.

%\section*{Acknowledgments}
%This was supported in part by......

\newpage
%Bibliography
\bibliographystyle{unsrt}  
\bibliography{references}

\end{document}